\documentclass{article}
\usepackage{amsmath,epsfig}
\usepackage{bm,amsfonts}
\usepackage[preprint]{spconfa4}

\copyrightnotice{\copyright 978-1-6654-3864-3/21/\$31.00 ©2021 IEEE }

\let\OLDthebibliography\thebibliography
\renewcommand\thebibliography[1]{
  \OLDthebibliography{#1}
  \setlength{\parskip}{0pt}
  \setlength{\itemsep}{0pt plus 0.3ex}
}

\begin{document}\sloppy

\def\x{{\mathbf x}}
\def\L{{\cal L}}

\title{EXPLORING EXPLICIT AND IMPLICIT VISUAL RELATIONSHIPS FOR IMAGE CAPTIONING}
%
\name{Zeliang Song\textsuperscript{\rm 1,2}, Xiaofei Zhou\textsuperscript{\rm 1,2} }
\address {
	\textsuperscript{\rm 1}Institute of Information Engineering, Chinese Academy of Sciences, Beijing, China \\
	\textsuperscript{\rm 2}School of Cyber Security, University of Chinese Academy of Sciences, Beijing, China \\
	\{songzeliang, zhouxiaofei\}@iie.ac.cn
}

\maketitle
\begin{abstract}
	Image captioning is one of the most challenging tasks in AI, which aims to automatically generate textual sentences for an image. Recent methods for image captioning follow encoder-decoder framework that transforms the sequence of salient regions in an image into natural language descriptions. However, these models usually lack the comprehensive understanding of the contextual interactions reflected on various visual relationships between objects. In this paper, we explore explicit and implicit visual relationships to enrich region-level representations for image captioning. Explicitly, we build semantic graph over object pairs and exploit gated graph convolutional networks (Gated GCN) to selectively aggregate local neighbors' information. Implicitly, we draw global interactions among the detected objects through region-based bidirectional encoder representations from transformers (Region BERT) without extra relational annotations. To evaluate the effectiveness and superiority of our proposed method, we conduct extensive experiments on Microsoft COCO benchmark and achieve remarkable improvements compared with strong baselines.
\end{abstract}

\section{Introduction}
Image captioning is a challenging task that connects computer vision and natural language processing. It aims to automatically describe the visual contents in an image with  natural languages and has attracted increasing interest in the field of artificial intelligence. This technique can be helpful for blind navigation and early education, and can be widely applied to cross-modal retrieval \cite{Cross} and human-robot interactions \cite{Robot}.

In the past few years, many encoder-decoder based models \cite{SAT,GLSTM} have been proposed to solve image captioning, where encoder employs convolutional neural networks (CNN) to extract semantic embeddings from the image and decoder utilizes recurrent neural networks (RNN) to generate captions. Inspired from machine translation and object detection, this framework has achieved striking advances through the introduction of visual attention mechanism \cite{SAAT} and reinforcement learning (RL) based training algorithm \cite{SCST}. Recently, attentive models have been improved by replacing attention over a uniform grid of features with attention over salient image region representations \cite{BUTD,GCNLSTM,AOA}. Regardless of these advancements, the researches on how to synthetically understand associations between objects for describing an image are not adequately explored. 

In the literature \cite{ROC}, it is well known that utilizing contextual information is essential for rich semantic understanding of complex visual scenes. A natural way to learn the contextual representation of an object is in terms of its visual relationships that characterize the interactions or associations with other objects in an image. On the one hand, some researches \cite{GCNLSTM,SGAE} exploit scene graphs constructed by the predicted visual relationships between objects to acquire context-aware region representations.  On the other hand, many studies \cite{AOA,M2,NGA} employ self-attention based transformer to adaptively  learn relevent weights between each object pairs. However, these methods have the following issues: (1) graph based approaches depend highly on the pre-trained visual relationship extractor, which may cause relationship prediction errors and omissions, resulting in  damaging performance of image captioning. (2) transformer based approaches model object interactions without explicit relation guidance, which will make it difficult for the decoder to understand the enhanced region representations.



To address the above problems, we creatively explore explicit and implicit visual relationships to learn context-aware region representations for image captioning in this paper.  Explicitly, we build a reduced semantic graph and design Gated GCN to  selectively focus on important edges. Implicitly, we completely depending on the region-based bidirectional encoder representations from transformers  (Region BERT) to draw the global dependencies between detected objects without explicit relational annotations. Subsequently, the two kinds of enriched region representations are fed into a language decoder together for generating captions. Instead of simply incorporating explicit and implicit visual features, we propose a Dynamic Mixture Attention (DMA) module to selectively focus on different types of features in a channel-wise manner. To validate the effectiveness of our proposal, we conduct extensive experiments and analysis on challenging Microsoft COCO benchmark and achieve significant improvements compared with strong baselines.

The main contributions are summarized as follows:
\begin{itemize}
	\item We explicitly construct semantic graph over object pairs filtered by their relative spatial relations and then exploit Gated GCN to selectively aggregate local neighbors' information, resulting in explicit enchanced region features.
	\item We devise two tasks to pre-train Region BERT which is used to implicitly draw global contextual associations among objects through self-attention. The Region BERT can provide implicit enhanced region features for language decoder.
	\item We present a Dynamic Mixture Attention module to better leverage explicit and implicit features for decoding stage. DMA separately performs attention mechanism between two kinds of features and hidden state, and then mixes the attention results according to  relevance with hidden state.
\end{itemize}

\section{Related Work}

\subsection{Image captioning}
Inspired by the recent advances in neural machine translation, current image captioning models usually follow the typical encoder-decoder framework \cite{SAT}, i.e., they use a CNN to encode an image and a RNN decoder to generate captions. Xu et al. \cite{SAAT} proposed soft and hard attention mechanisms to focus on different image regions when generating different words. Lu et al. \cite{Knowing} improved this work through introducing a visual sentinel to allow attention mechanism selectively attend to visual and language features. Rennie et al. \cite{SCST} presented SCST algorithm, which is a form of the popular REINFORCE algorithm that utilizes the output of its own test-time inference algorithm to normalize the rewards it experiences. SCST has become the most popular training method, because it only needs one additional forward propagation.

\subsection{Contextual region representations}
With the release of the Visual Genome \cite{VG} dataset, most approaches employ faster R-CNN pre-trained on this dataset to extract more fine-grained region representations for an image. Anderson et al. \cite{BUTD} enabled attention to be calculated at the level of objects through a combined bottom-up and top-down attention mechanism. Yao et al. \cite{GCNLSTM} integrated semantic and spatial relationships between objects into long short term memeory (LSTM) decoder via graph convolutional networks (GCN).  Yang et al. \cite{SGAE} used a directed graph where an object node is connected by adjective nodes and relationship nodes to represent complex hierarchical information of both image and sentence. Huang et al. \cite{AOA} 
proposed the Attention on Attention (AoA) module, an extension to the conventional attention mechanism, to better model relationships among different objects in the image. Cornia et al. \cite{M2} devised memory augmented transformer that models the  relationships between image regions in a multi-level fashion exploiting learned a priori knowledge. Guo et al. \cite{NGA} proposed Normalized Self-Attention that brings the benefits of normalization inside self-attention, and a class of Geometry-aware Self-Attention that extends self-attention to consider the relative geometry relations between the objects. In this paper, we propose to explore the complementary characteristic of graph based and transformer based visual features for improving image captioning.

\begin{figure*}[t]
	\begin{center}
		\includegraphics[width=0.83\linewidth]{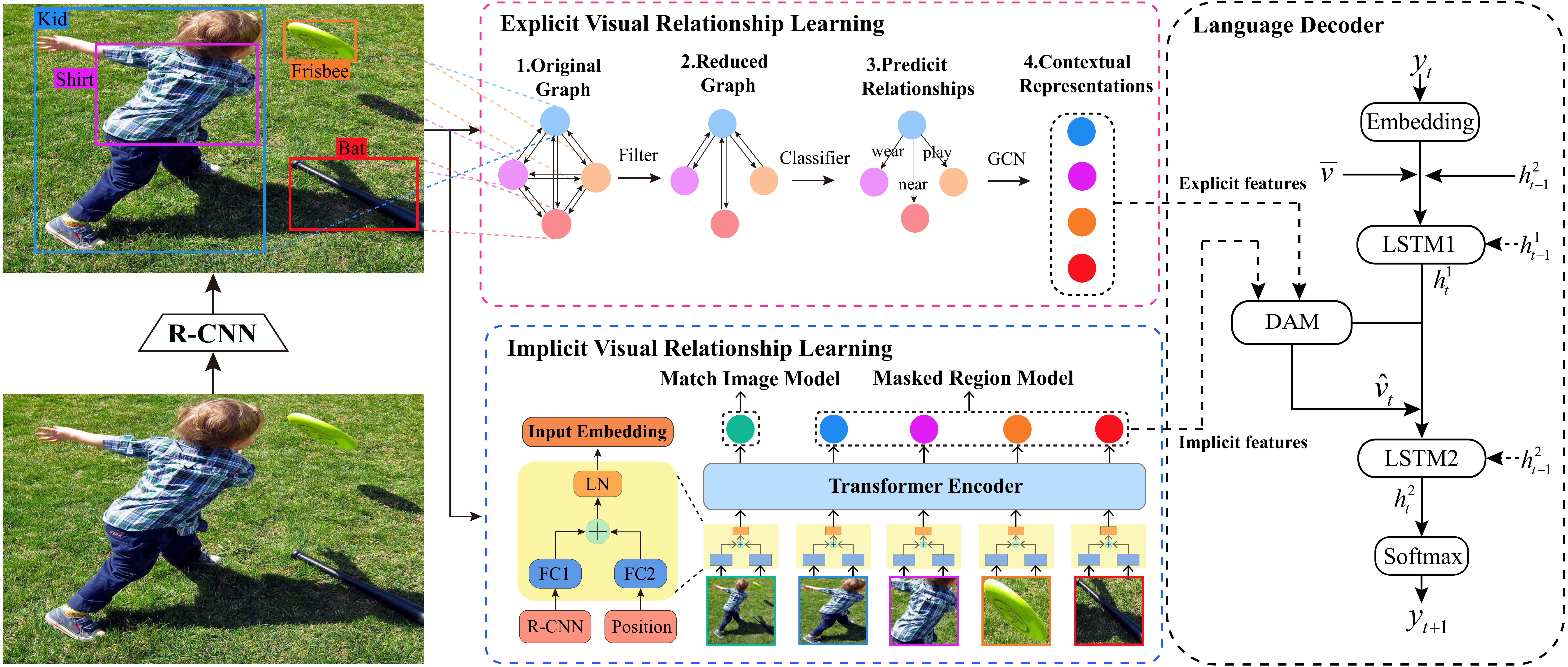}
	\end{center}
	\caption{An illustration of our proposed method. Firstly, it employs faster R-CNN to extract visual features at the level of objects. Secondly, it explores explicit and implicit visual relationships among objects  to achieve context-aware region representations. Finally, these enriched region features are fed into attention based language decoder to generate captions.}
	\label{workflow}
\end{figure*}  

\section{Approach}
In this section, we will introduce how to explicitly and implicitly explore visual relationships for image captioning. We firstly utilize faster R-CNN to encode an image into a set of $k$ region features $\mathcal{V}=\{\bm{v}_1,\bm{v}_2,...,\bm{v}_k\}, \bm{v}_i \in \mathbb{R}^{v}$. Then we respectively explore explicit and implicit visual relationships among objects via Gated GCN and Region BERT, resulting in enhanced explicit features $\hat{\mathcal{V}}_x$ and implicit features $\hat{\mathcal{V}}_m$. Finally, we feed these contextual representations into Dynamic Mixture Attention (DMA) module for generating captions. An overview of our method is illustrated in Figure \ref{workflow}. 

\subsection{Exploring explicit visual relationships}

Given the detected region features $\mathcal{V}$, the original graph is defined as a directed complete graph $\mathcal{G}_o=\{\mathcal{V}, \mathcal{E}_o\}$,  where $\mathcal{V}$ and $\mathcal{E}_o$ are the sets of nodes and edges. Note that each image region $\bm{v}_i$ is one vertex and each edge represents a type of relationship. Based on the original graph, we explore explicit visual relationship  via a Gated GCN. 

\subsubsection{Graph construction} 
In order to accelerate the construction of the semantic graph and improve the accurate of visual relationship classifier, we leverage an edge filter to delete edges between object pairs in $\mathcal{G}_o$ that have a large spatial relative distance, resulting a reduced graph $\mathcal{G}_r$. To be specific, given two regions $\bm{v}_i$ and $\bm{v}_j$, their locations are denoted as $(x_i, y_i)$ and $(x_j, y_j)$, that are the centroid coordinates of the bounding box for each region. Thus we can compute Intersection over Union (IoU) and relative distance $d_{ij}=\sqrt{(x_i-x_j)^2+(y_i-y_j)^2}$ between them. The edges ($\bm{e}_{ij}$ and $\bm{e}_{ji}$) are removed from $\mathcal{G}_o$ if and only if $IoU=0$ and $d_{ij}>l_{ij}$, where $l_{ij}$ is length of the longest edge in two bounding boxes.
In our implementation, we take MOTIFS detector \cite{MOTIFS} as our relationship classifier to predict the type of relation for each directional edge in $\mathcal{G}_r$. Note that we keep the edges with higher probability to make sure there is at most one relationship between each object pairs.  Consequently, a semantic graph $\mathcal{G}_s=\{\mathcal{V}, \mathcal{E}_s\}$ is completed.

\subsubsection{Gated GCN} 
With the constructed semantic graph $\mathcal{G}_s$, we integrate the predicted visual relationships into contextual region representations learning via a Gated GCN. To make our GCN sensitive to both directionality and labels, each region vertex $\bm{v}_i$ is encoded as follows: 
\begin{equation}\label{GCN}
\bm{v}_{i}^{\prime}=ReLu(\sum_{\bm{v}_j \in \mathcal{N}(\bm{v}_i)} \bm{W}_{dir(i,j)} \bm{v}_j + \bm{b}_{lab(i,j)})
\end{equation}
where $\mathcal{N}(\bm{v}_i)$ represents the set of neighbors of $\bm{v}_i$ and also includes $\bm{v}_i$ itself; $\bm{W}_{dir(i,j)} \in \mathbb{R}^{d \times v}$ and $dir(i,j)$ indicates the direction (\emph{i.e.}, 1 for $\bm{v}_i$-to-$\bm{v}_j$, 2 for $\bm{v}_j$-to-$\bm{v}_i$, and 3 for $\bm{v}_i$-to-$\bm{v}_i$); $\bm{b}_{lab(i,j)} \in \mathbb{R}^d$ and $lab(i,j)$ represents the type of edge. Considering the importance of each edge are different, we incorporate an edge-wise gate mechanism into GCN:
\begin{gather}\label{Gated GCN}
\bm{v}_{i}^{\prime}=ReLu(\sum_{\bm{v}_j \in \mathcal{N}(\bm{v}_i)} g_{ij}(\bm{W}_{dir(i,j)} \bm{v}_j + \bm{b}_{lab(i,j)}) ) \\
g_{ij} \propto \bm{w}_g^T tanh(\bm{W}_{a}\bm{v}_{sub(i,j)} + \bm{W}_b \bm{v}_{obj(i,j)})
\end{gather}
where $g_{ij}$ is a normalized attention weight; $\bm{W}_a \in \mathbb{R}^{d \times v}$, $\bm{W}_b \in \mathbb{R}^{d \times v}$, and $\bm{w}_g \in \mathbb{R}^{d}$ are learned parameters; $sub(i,j)$ and $obj(i,j)$ denote the subject and object respectively.

\subsection{Exploring Implicit Visual Relationships}
In this section, we will introduce how to implement implicit visual relationship learning through Region BERT.  Then we employ a transformer encoder to draw global interactions among objects. Next, we pre-train our Region BERT on two tasks, i.e., Match Image Modeling (MIM) and Masked Region Modeling (MRM). 

\subsubsection{Region BERT} 
As shown in Figure \ref{workflow}, the inputs of Region BERT includes a global image feature and a set of region features. The input embedding consists of the pooled visual feature $\bm{v}_i$ and corresponding positional feature $\bm{p}_i$, where $\bm{p}_i$ is a 6-dimensional vector (Normalized top/bottom/left/right coordinates, width, height). Both visual and positional features are respectively projected into the same embedding space via a fully-connected (FC) layer. The final input embedding for each location is layer-normalized sum of the two FC outputs. Then these input embeddings are passed through a transformer encoder, resulting in contextual region representations.

\subsubsection{Match image modeling} 
In MIM, when choosing the global visual feature $\bm{v}_g$, 50\% of the time $\bm{v}_g$ is the actual feature, 50\% of the time it is a random image feature selected from the dataset. We apply an FC layer upon the transformer output corresponding to $\bm{v}_g$, scoring how well global image and salient regions are matched. Then our Region BERT is trained by minimizing the following loss function:
\begin{equation}\label{MIM}
\begin{split}
\mathcal{L}_{MIM} = -\mathbb{E}_{(\bm{v}_g, \mathcal{V}) \sim \mathcal{D}} [&y \log P(\bm{v}_g, \mathcal{V}) \\
&+(1-y) \log (1-P(\bm{v}_g, \mathcal{V}))]
\end{split}
\end{equation}
where $y\in \{0,1\}$ indicates if the pair $(\bm{v}_g, \mathcal{V})$ is a match.

\subsubsection{Masked region modeling}
In MRM,  we sample image regions from $\mathcal{V}$ and mask their visual features with a probability of 10\%. 
If region $\bm{v}_i$ is chosen, we replace it with (1) zeros 80\% of the time (2) a random region feature 10\% of the time (3) the unchanged feature 10\% of the time. We apply an FC layer upon the transformer output to reconstruct the selected region based on masked region features $\mathcal{V}_{mask}$ by minimizing the following loss function:
\begin{equation}
\mathcal{L}_{MRM} = \mathbb{E}_{(\bm{v}_g, \mathcal{V}) \sim \mathcal{D}^{\prime}} [\sum_{i} ||f_i(\bm{v}_g, \mathcal{V}_{mask}) - \bm{v}_i||^2_2]
\end{equation}
where $f_i(\bm{v}_g, \mathcal{V})$ is the reconstructed region feature with respect to masked region $\bm{v}_i$.

\subsection{Dynamic mixture attention}
After obtaining the explicit features $\hat{\mathcal{V}}_x=\{\bm{x}_{1},\bm{x}_{2},...,\bm{x}_{k}\}$ from Gated GCN and implicit features $\hat{\mathcal{V}}_m=\{\bm{m}_{1},\bm{m}_{2},...,\bm{m}_{k}\}$ from Region BERT, DMA first processes them with attention mechanism  based on current hidden state $h_t$:
\begin{gather}
\hat{\bm{x}}_t = f_{att1}(\hat{\mathcal{V}}_x, \bm{h}_t) \\
\hat{\bm{m}}_t = f_{att2}(\hat{\mathcal{V}}_m, \bm{h}_t) 
\end{gather}
where the attention operation $f_{att}$ is the same with \cite{BUTD}.
The DMA calculates a gated information in a manner of channel-wise to dynamically control the attention level of each kind of feature. The attention results $\hat{\bm{x}}_t$ and $\hat{\bm{m}}_t$ are concatenated together with hidden state $\bm{h}_t$, and then the concatenated vector is used to generate an attention gate $\bm{g}_t$:
\begin{equation}
\bm{g}_t = \sigma(\bm{W}_g[\hat{\bm{x}}_t;\hat{\bm{m}}_t;\bm{h}_t])
\end{equation}
where $\bm{W}_g \in \mathbb{R}^{d\times3d}$, and $\sigma(\cdot)$ is the sigmoid activation function. The final output of DMA is given by:
\begin{equation}
\hat{\bm{v}}_t = \bm{g}_t \odot \hat{\bm{x}}_t + (1-\bm{g}_t) \odot \hat{\bm{m}}_t 
\end{equation}
where $\odot$ denotes  element-wise multiplication.

\begin{table}[t]
	\begin{center}
		\caption{Performance comparisons of various methods on MSCOCO Karpathy split.  The metrics: B-4, M, R, C, S denote BLEU-4, METEOR, ROUGE-L, CIDEr-D, SPICE.}
		\label{offline}
		\scalebox{0.9}{\begin{tabular}{|l|c|c|c|c|c|}
				\hline
				Models   & B-4 & M & R & C & S\\
				\hline
				
				Up-Down \cite{BUTD}   & 36.3 & 27.7 & 56.9 & 120.1 & 21.4 \\
				SGAE \cite{SGAE}  & 38.4 & 28.4 & 58.6 & 127.8 & 22.1\\
				GCN-LSTM \cite{GCNLSTM}  & 38.3 & 28.6 & 58.5 & 128.7 & 22.1\\
				AoANet \cite{AOA}  & 39.1 & 29.2 & 58.8 & 129.8 & 22.4\\
				$\mathcal{M}^2$ Transformer \cite{M2} & 39.1 & 29.2& 58.6 & 131.2 & 22.6 \\
				\hline
				\textbf{EIVRN}$_{LSTM}$       & 38.9 & 28.9 & 58.7 & 129.6 & 22.4       \\
				\textbf{EIVRN}$_{Trans}$  & \textbf{39.4} & \textbf{29.3} & \textbf{59.1} & \textbf{131.9} & \textbf{22.8}\\
				\hline
		\end{tabular}}
	\end{center}
\end{table}

\subsection{Training}
Given the ground truth caption $\mathcal{S}^*$  for $\mathcal{I}$, we pre-train our captioning model by minimizing the following cross-entropy loss:
\begin{equation}\label{cross-entropy}
\mathcal{L}_{XE}(\bm{\theta}) = -\log P(S^*|\mathcal{V})
\end{equation}
where $\bm{\theta}$ denotes model parameters. However, minimizing cross-entropy inevitably suffers from exposure bias problem. Accordingly, to further enhance image captioning by addressing such issue, we can directly optimize our captioning model by maximizing the RL-based expected reward:
\begin{equation}\label{RL}
\mathcal{L}_{RL}(\bm{\theta}) = \mathbb{E}_{\mathcal{S}^s \sim P_{\bm{\theta}} } [r(\mathcal{S}^s, \mathcal{S}^*)]
\end{equation}
where $r(\cdot, \cdot)$ is a sentence-level metric computed between the sampled sentence $\mathcal{S}^s$ and the ground truth sentence $\mathcal{S}^*$, \emph{e.g.}, the CIDEr-D \cite{CIDEr} metric.

\section{Experiments}

\subsection{Dataset, metrics and setting}
\subsubsection{Dataset}
We evaluate our proposed method on the most popular Microsoft COCO 2014 captions dataset. Each image in Microsoft COCO has at least 5 human-annotated descriptions. In this paper, we employ widely-used Karpathy split (https://github.com/karpathy/neuraltalk) which contains 113,287/5,000/5,000 images for training/validation/testing. We truncate all the sentences in training set to ensure that any sentence does not exceed 16 characters. We follow standard practice \cite{BUTD} and perform only minimal text pre-processing: tokenizing on white space, converting all sentences to lower case, and filtering words which occur less than 5 times, resulting in a vocabulary of 10,369 words. 
\subsubsection{Metrics}
To evaluate quantitative performance, we use five standard automatic evaluation metrics, including BLEU \cite{BLEU}, ROUGE-L \cite{ROUGE}, METEOR \cite{METEOR}, CIDEr-D \cite{CIDEr} and SPICE \cite{SPICE}. All the metrics are computed with the publicly released evaluation tool (https://github.com/tylin/coco-caption).
\subsubsection{Setting}
We implement our method on a LSTM-based decoder \cite{BUTD} and a Transformer-based decoder, and we call them EIVRN$_{LSTM}$ and EIVRN$_{Trans}$. The attention hidden size of GCN and transformer encoder is 512, and we stack 6 transformer blocks. The setting of LSTM decoder is the same with \cite{BUTD} for fair comparisons. The word embedding size and hidden size of transformer decoder are 512. We use beam search with a beam size of 3 when validating and testing. We first pre-train Region BERT encoder on MIM and MRM tasks for 20 epochs.
Then we pre-train our captioning model under cross-entropy loss for 30 epochs with a batch size of 10. We employ ADAM optimizer with an initial learning rate of $2e^{-4}$, and momentum of 0.9 and 0.999. We evaluate our model on the validation set at every epoch and select the model with highest CIDEr-D score as the initialization for self-critical training. The learning rate for SCST is initialized as  $1e^{-4}$ and decays by 0.1 exponentially  every 50 epoches.

\subsection{Analysis}

\subsubsection{Quantitative analysis}
We compare our models (\textbf{EIVRN}$_{LSTM}$ and \textbf{EIVRN}$_{Trans}$) with the following state-of-the-art methods: \textbf{Up-Down} \cite{BUTD}, \textbf{SGAE} \cite{SGAE}, \textbf{GCN-LSTM} \cite{GCNLSTM}, \textbf{AoANet} \cite{AOA}, \textbf{M$^2$ Transformer} \cite{M2}. The performance comparisons on Microsoft COCO Karpathy split are shown in Table \ref{offline}. From this table, we can see that EIVRN$_{LSTM}$ achieves competitive performance and EIVRN$_{Trans}$ significantly outperforms other methods. Specially, EIVRN$_{Trans}$ achieves 39.4/29.3/59.1/131.9/22.8 on B-4/M/R/C/S respectively. Focusing on the important CIDEr-D metric, EIVRN$_{Trans}$ outperforms advanced $M^2$ Transformer by 0.7. The results of comparison demonstrate the effectiveness  and superiority of our proposed methods, especially on those more convincing evaluation metrics such as CIDEr and SPICE. 

\begin{table}[t]
	\begin{center}
		\caption{Performance comparisons of different variant models of EIVRN on Karpathy split.}
		\label{ablative}
		\scalebox{0.9}{\begin{tabular}{|l|c|c|c|c|c|}
				\hline
				Models   & B-4 & M & R & C & S\\
				\hline
				
				EVRN                  & 37.9 & 28.2 & 58.0 & 127.1 & 21.8 \\
				IVRN$_{BASE}$ & 38.2 & 28.3 & 58.2& 127.9 & 22.0 \\
				IVRN$_{MRM}$  & 38.3 & 28.3 & 58.3 & 128.1 & 22.1\\
				IVRN$_{MIM}$   & 38.5 & 28.5 & 58.4 & 128.5 & 22.1\\
				IVRN                   &38.5 & 28.6 & 58.5 & 128.8& 22.2\\
				EIVRN$_{ADD}$  &38.7 & 28.7 & 58.6 & 129.2 & 22.3     \\
				\hline
				\textbf{EIVRN}   & \textbf{38.9} & \textbf{28.9} & \textbf{58.7} & \textbf{129.6} & \textbf{22.4}       \\
				\hline
		\end{tabular}}
	\end{center}
\end{table}
\subsubsection{Ablative analysis}
In this section, we extensively explore different structures and settings of our method to gain insights about how and why it works. We select EIVRN$_{LSTM}$ model for ablation studies, and we call it EIVRN next for simplicity. We devise the following variant models: (1) \textbf{EVRN} only uses explicit features for decoding; (2) \textbf{IVRN} only uses implicit features for decoding; (3) \textbf{IVRN}$_{BASE}$ is IVRN without pre-training Region BERT; (4) \textbf{IVRN}$_{MRM}$ only pre-trains Region BERT on MRM task; (5) \textbf{IVRN}$_{MIM}$ only pre-trains Region BERT on MIM task; (6) \textbf{EIVRN}$_{ADD}$ simply adds the explicit and implicit features for decoding. 

The ablative results are shown in Table \ref{ablative}, from which we can observe that: (1) IVRN performs better than EVRN, this is due to IVRN can be deeper with the residual connection and layer normalization.
(2) Pre-training the Region BERT on MIM and MRM tasks can provide an optimistic initialization, and pre-training with MIM task can achieve  slightly better results. (3) After jointly pre-training on  MRM and MIM, IVRN outperforms other variant models of it. (4) EIVRN$_{ADD}$ outperforms EVRN and IVRN, which suggests that simply combining explicit and implicit features can also bring improvements. (5) EIVRN is better than EIVRN$_{ADD}$, which demonstrates that DMA module can help language decoder leverage explicit and implicit features more effectively. The above observations validate the effectiveness of our proposed method.

\begin{figure}[t]
	\begin{center}
		\includegraphics[width=0.82\linewidth]{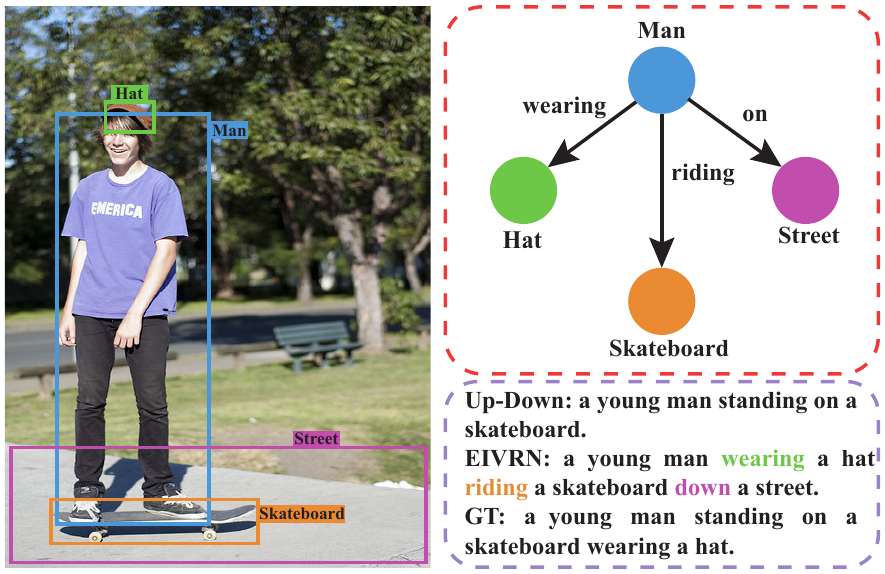}
	\end{center}
	\caption{This figure shows an example of the constructed semantic graph with GT caption, and captions from EIVRN and Up-Down.}
	\label{graph}
\end{figure} 
\subsubsection{Qualitative analysis}
Figure \ref{graph} shows an visualization of the constructed semantic graph with ground truth (GT) sentences and sentences generated by EIVRN and Up-Down. From which we can see that the detected visual relationships among objects can help our model generate captions with more interactive words, \emph{e.g.}, ``wearing'', ``riding'' and ``down''. Specially, EIVRN even can generate caption with interactive word which doesn't exist in ground truth caption, \emph{i.e.}, the relation word ``down''  between objects ``man'' and ``street'', which further demonstrates the effectiveness of our method.

\section{Conclusion}
In this paper, we creatively explore explicit and implicit visual relationships that draw contextual interactions between objects to enrich region representations for image captioning. On the one hand, we explicitly predict relationships between filtered object pairs and use Gated GCN to selectively aggregate local neighbors' information. On the other hand, we implicitly learn global interactions among objects through Region BERT pre-trained on designed MIM and MRM tasks. We evaluate our proposed method on Microsoft COCO benchmark, and we achieve remarkable improvements over strong baselines.

\bibliographystyle{IEEEbib}
\bibliography{icme2021template}

\end{document}